\documentclass[10pt,twocolumn,letterpaper]{article}

\usepackage[pagenumbers]{iccv}     
\usepackage{etoc}
\usepackage{minitoc}
\usepackage{bm}

\usepackage{extarrows}

\definecolor{cvprblue}{rgb}{0.21,0.49,0.74}
\usepackage[pagebackref,breaklinks,colorlinks,allcolors=cvprblue]{hyperref}
\newcommand{\ourwork}{V$^2$Edit\xspace}
\newcommand\themodel{\ourwork} 
\newcommand{\updated}[1]{#1} 

\def\paperID{ArXiV} 
\def\confName{ArXiV}
\def\confYear{2025}

\title{\themodel: Versatile Video Diffusion Editor for Videos and 3D Scenes}

\author{Yanming Zhang$^{1}$$^\dagger$ \qquad Jun-Kun Chen$^{2}$$^\dagger$ \qquad Jipeng Lyu$^{2}$\qquad Yu-Xiong Wang$^{2}$ \vspace{0.1em} \\ 
    $^1$Zhejiang University \qquad $^2$University of Illinois Urbana-Champaign \qquad $^\dagger$Equal Contribution \vspace{0.1em}\\
    {\tt \hspace{0mm}yanmingzhang@zju.edu.cn \qquad{\tt \{junkun3,jipeng2,yxw\}@illinois.edu}} \vspace{0.2em}\\ \href{https://immortalco.github.io/V2Edit/}{\textbf{\texttt{immortalco.github.io/V2Edit}}}
}

\begin{document}

\twocolumn[{
\maketitle
\vspace{-8.0mm}
\renewcommand\twocolumn[1][]{#1}
    \centering
    
    \includegraphics[width=1.0\linewidth]{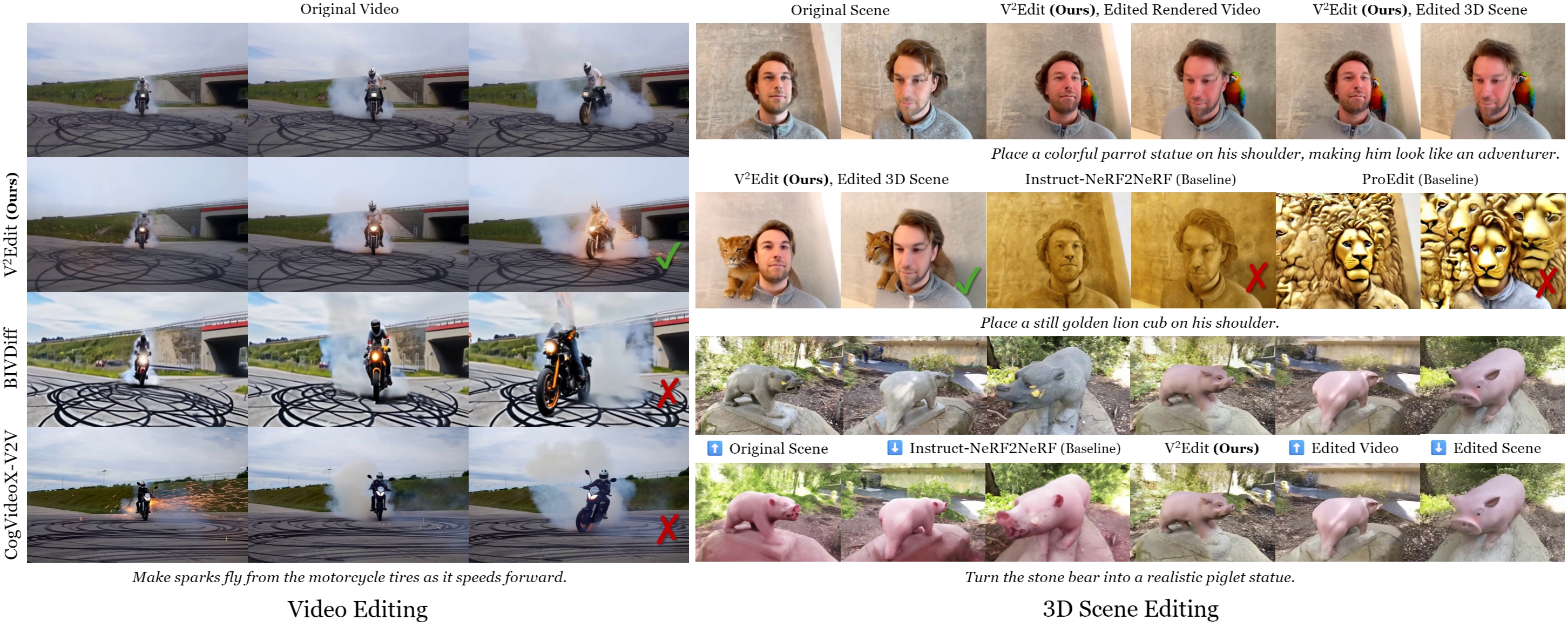}
    \vspace{-8mm}
    \captionof{figure}{Our \textbf{\themodel} is a versatile approach that supports \emph{training-free} instruction-guided editing for both videos and 3D scenes. \textbf{Left:} \themodel achieves high-quality editing satisfying both \emph{original content preservation} and \emph{editing instruction fulfillment} in video editing. \textbf{Right:} \themodel supports challenging 3D scene editing tasks involving \emph{significant geometric changes}, which baselines \cite{in2n,proedit} fail to achieve. } 
    
\label{fig:teaser}
    \vspace{2mm}
}]

\begin{abstract}
\vspace{-5mm}

\noindent This paper introduces \themodel{}, a novel training-free framework for instruction-guided video and 3D scene editing. Addressing the critical challenge of balancing original content preservation with editing task fulfillment, our approach employs a progressive strategy that decomposes complex editing tasks into a sequence of simpler subtasks. Each subtask is controlled through three key synergistic mechanisms: the initial noise, noise added at each denoising step, and cross-attention maps between text prompts and video content. This ensures robust preservation of original video elements while effectively applying the desired edits. Beyond its native video editing capability, we extend \themodel{} to 3D scene editing via a ``render-edit-reconstruct'' process, enabling high-quality, 3D-consistent edits even for tasks involving substantial geometric changes such as object insertion. Extensive experiments demonstrate that our \themodel{} achieves high-quality and successful edits across various challenging video editing tasks and complex 3D scene editing tasks, thereby establishing state-of-the-art performance in both domains.

\end{abstract}
\vspace{-3mm}    
\section{Introduction}
\label{sec:intro}
Video diffusion models have rapidly gained prominence in computer vision~\cite{svd,cogvideo,sora,sv3d,videoldm}, following the success of image-based diffusion generative models \cite{sd}. These models now enable the generation of high-resolution, high-fidelity videos from text descriptions. Meanwhile, instruction-guided video editing -- modifying existing videos through simple text instructions -- has become an emerging area of focus. Using a high-quality initial video allows for efficient creation of new video assets through targeted edits, rather than generating them from scratch.

Despite such progress, video editing remains under-explored due to the lack of large-scale paired video datasets essential for training end-to-end models. Traditional image-based methods~\cite{fatezero,stylizingvideo,shapeaware,i4d24d,slicedit} approach video editing by applying image editing techniques to individual frames, one at a time. The advent of video diffusion models has enabled \emph{training-free} video-based methods~\cite{bivdiff,videoshop} that leverage pre-trained video diffusion models without the need for additional training~\cite{sdedit,ddpminv}. However, these methods still face challenges, including temporal inconsistencies, as well as difficulties in handling fast-moving camera trajectories, complex motions, and significant temporal variations.

\updated{Our main observation is that} video editing tasks require simultaneously achieving two objectives: the edited video should both \emph{fulfill the editing instruction} and \emph{preserve the original content}. This ensures that only the targeted areas are modified, while all other components remain unchanged. Balancing these two aspects, however, presents a key challenge in video editing. The existing training-free models are often driven to produce videos that comply with the editing instructions but fail to preserve the original content, \updated{as well as requiring extensive hyperparameter tuning to achieve a balance.}

This paper proposes \themodel, a novel framework for versatile video editing that introduces effective and synergistic control mechanisms, robustly preserving the original content while remaining flexible to allow for the intended edits. To preserve the original video content during editing, \emph{our first key insight} is to systematically control the denoising process in video diffusion models from complementary perspectives: \textbf{(i)} the initial noise, \textbf{(ii)} the noise added at each denoising step, and \textbf{(iii)} the cross-attention maps between text prompts and the video.

Specifically, the noise addition process in diffusion models initially disrupts high-frequency details and later affects low-frequency information. By restricting the noise addition to the early steps and starting generation from this initial noise, we can preserve low-frequency features, such as the \emph{overall layout}. This observation further suggests that noise carries semantic information in diffusion-based generation. Therefore, using a noise scheduler that incrementally adds noise at each step can help transfer \emph{semantic} information from the original video to the edited one. Additionally, during the denoising process, the model’s cross-attention maps -- showing \emph{correspondences} between textual prompts and specific objects or regions -- can be explicitly leveraged to control the preservation of the original content.

Conceptually, these control mechanisms may need to be tailored to suit different editing tasks: mild edits can allow for stronger preservation of the original content, whereas significant edits may be compromised with too strict preservation control. To avoid such complexity, our \emph{second key insight} is to \emph{decompose} a complex editing task into a sequence of mild subtasks, progressively completing each subtask. For each subtask, it becomes easier to balance the original content preservation with editing sub-instruction fulfillment. More importantly, this can be achieved with a \emph{consistent control strategy} across subtasks derived from different editing tasks, rather than relying on a more complex, task-varying, \updated{or \emph{hyperparameter-sensitive} approach}.

Beyond its native video editing capability, our \themodel can also be seamlessly applied to 3D scene editing, making it a \emph{unified} editing solution. \updated{We propose a simple yet effective ``render-edit-reconstruct (RER)'' process to leverage video editing methods for 3D editing}, by first rendering a video of the scene along a fixed camera trajectory, editing the rendered video, and then reconstructing the scene from the edited video. The temporal consistency of the rendered video ensures strong 3D consistency in the reconstructed scene. In fact, 3D scene editing is a \emph{unique} capability of our \themodel, \updated{where existing video editing methods \cite{bivdiff,i4d24d,videoshop,slicedit,csd} struggle to} render videos with large-scale camera motions and significant temporal variations. 

In a variety of challenging video and 3D scene editing tasks, our \themodel achieves high-quality results, as shown in Fig.~\ref{fig:teaser}. For video editing, our method handles more complex scenarios, including longer videos, faster-moving camera trajectories, and greater temporal variations. In 3D scene editing, \themodel notably supports significant geometric changes, such as object insertion, which previous 3D scene editing methods \cite{in2n,en2n,proedit} have not been able to accommodate. Additionally, our approach enables efficient video editing without the need for time-consuming, iterative per-view adjustments, ensuring rapid convergence.

\textbf{Our contributions} are threefold. \textbf{(i)} We propose \themodel, a simple yet versatile framework for training-free, instruction-guided video and 3D scene editing. \textbf{(ii)} We introduce synergistic mechanisms that systematically control the denoising process in video diffusion and enable progressive editing, effectively balancing the preservation of original video content with the fulfillment of editing instructions, all within a unified framework for diverse editing tasks. \textbf{(iii)} \themodel consistently achieves high-quality, successful edits across various video and 3D scene editing tasks, including those previously unsolvable by existing methods, thereby establishing state-of-the-art performance in both domains.
\section{Related Work}
\label{sec:related}

\noindent\textbf{Video Diffusion Models.} The success of diffusion models in image generation has been extended to video generation~\cite{vdm}. Early approaches \cite{vdm,dpmvg,makeavideo,imagenvideo,videoldm} design the video diffusion model based on the UNets of image diffusion models, to support the 3D-shaped inputs for videos. To save memory and compute, instead of directly lifting the convolutional layers and attention layers from 2D to 3D, they keep the existing 2D layers to be applied individually to each frame, while inserting temporal convolutional and attention layers. This decomposes the computation of spatial and temporal components of videos, and also makes it possible to extend pre-trained image diffusion models by only tuning the temporal generation capability through fine-tuning \cite{makeavideo,imagenvideo,videoldm}. Later, Stable Video Diffusion (SVD) \cite{svd} scales up the video diffusion models for high-resolution, high-quality video generation through careful data selection and multi-stage training, and also extends to the generation of 3D \cite{sv3d} and 4D \cite{sv4d} contents. The release of SORA \cite{sora} has lit a new way to scale up video diffusion models with diffusion transformers (DiTs). Instead of applying downsampling and decomposed attention layers in UNets, DiTs directly turn the whole video (or video latents) into a sequence of patches, and apply a full 3D attention within all the patches. Inspired by this, CogVideoX \cite{cogvideox} is proposed upon its previous effort CogVideo \cite{cogvideo}, using DiT-based video diffusion models and significantly improving the video length, resolution, and generation quality.

\noindent\textbf{Video Editing.} \updated{Due to the lack of paired training data for video editing methods -- \ie, triples of ``editing instruction, original video, and edited video'' -- most existing video editing approaches are training-free.} Traditional video editing methods \cite{fatezero,stylizingvideo,shapeaware,i4d24d,slicedit} are image-based methods, which rely on an underlying model or method with image editing capability and introduce other add-ons to control the consistency. For example, FateZero \cite{fatezero}, Tune-A-Video \cite{tuneavideo}, and Instruct 4D-to-4D \cite{i4d24d} use 2D diffusion models to edit videos by zero-shot extending the standard spatial attention layers in the UNets into spatial-temporal attention layers to account for the first and the previous frame in the generation. After the emergence of video diffusion models, there are also several efforts that edit videos by utilizing the generation and smoothness capability of video diffusion models. BIVDiff \cite{bivdiff} uses a pre-trained video diffusion model to refine the temporal-inconsistent per-frame edited images into a smooth, temporal-consistent video. VideoShop \cite{videoshop} takes the edited first frame of the video as input along with the video and propagates the editing operations through the following frames. CogVideoX-V2V \cite{cogvideox} applies SDEdit \cite{sdedit} to edit videos using the generation capability. However, unlike our \themodel, all these methods struggle to perform aggressive editing on complex scenarios, like fast-moving cameras, changing backgrounds and contents, and geometry or motion changes. \updated{Many training-free methods require task-specific hyperparameter tuning to manually balance editing fulfillment and original preservation, while our hyperparameter-tuning-free \themodel achieves a consistent preservation strategy across all editing tasks.}

\noindent\textbf{Diffusion-Based 3D Scene Editing.} For instruction-guided 3D scene editing, the traditional way is to distill the editing signals from a 2D diffusion applied on each view to the 3D scene with score distillation sampling (SDS) \cite{dreamfusion}. Instruct-NeRF2NeRF \cite{in2n} is the first paper in this direction, which applies an SDS-equivalent iterative dataset update to iteratively update the dataset of edited views to train the NeRF. There are also many works in this direction, aiming to improve the efficiency \cite{en2n}, distillation method \cite{pds,csd}, 3D consistency \cite{consistdreamer,proedit}, or extension to 4D \cite{i4d24d}. Video diffusion models show a natural and straightforward way to replace the image diffusion model with a video diffusion model, to edit a rendered video of the scene directly. As the temporal consistency of the edited rendered video is a strong prerequisite of 3D consistency of the 3D scene, this could potentially significantly reduce the difficulty of maintaining 3D consistency. However, the rendered videos of a 3D scene should cover all possible viewpoints of the scene, which requires the content to change a lot to cover the whole scene, and the camera trajectory covers all possible viewpoints. All these make such video editing a very challenging case in video editing tasks. To our knowledge, no existing work utilizes video diffusion models to edit 3D or 4D scenes.
\section{Methodology}
\label{sec:method}

\begin{figure*}[t!]
\centering
\centerline{\includegraphics[width=1\linewidth]{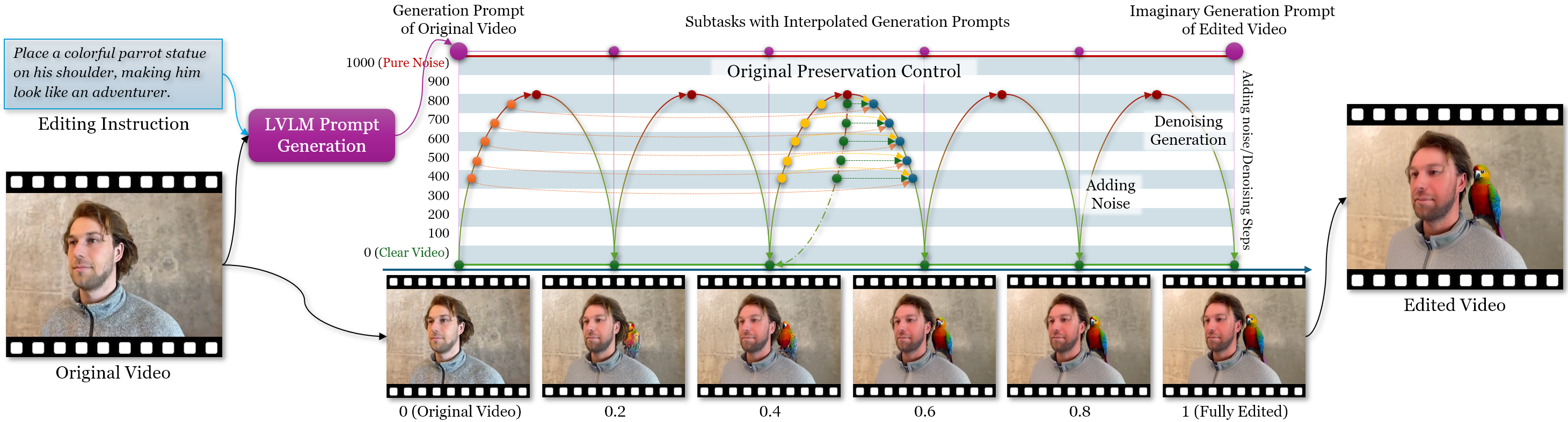}}
\vspace{-3mm}
\caption{\textbf{Our \themodel framework} features progressive editing. Given an editing instruction and the original video, a large vision-language model (LVLM) \cite{gpt4o} generates prompts for both the original and edited videos. These prompts are interpolated to create a sequence of subtasks, which are executed progressively in our framework.}
\vspace{-6.5mm}
\label{fig:method:pipeline}
\end{figure*}

In \themodel{}, we leverage pre-trained video diffusion models as the foundation for versatile video editing without requiring specific training on paired datasets. Our framework, illustrated in Fig.~\ref{fig:method:pipeline}, employs a progressive editing process that decomposes complex editing tasks into a sequence of simpler subtasks. To preserve the original video content while ensuring high-quality edits, we implement a training-free preservation control mechanism that systematically manages three key aspects of the diffusion process: \textbf{(i)} the initial noise, \textbf{(ii)} the noise added at each denoising step, and \textbf{(iii)} the cross-attention maps between text prompts and video content. This approach ensures that the original elements of the video are robustly maintained while effectively applying the intended modifications, \updated{through a consistent preservation control strategy without hyperparameter-tuning.}

\subsection{Prompt Generation}
\label{sec:method:prompt-gen}
We leverage large vision-language models (LVLMs)~\cite{gpt4o} to convert editing instructions into two descriptive prompts: one for the original video and another for the edited video. This is essential because most text-to-video diffusion models require prompts that describe the video content itself. By generating these tailored prompts, our framework ensures that the underlying diffusion model can effectively perform instruction-guided editing while maintaining the structure and integrity of the original video content.

\subsection{Original Preservation Control}

\begin{figure}[t!]
\centering
\centerline{\includegraphics[width=1\linewidth]{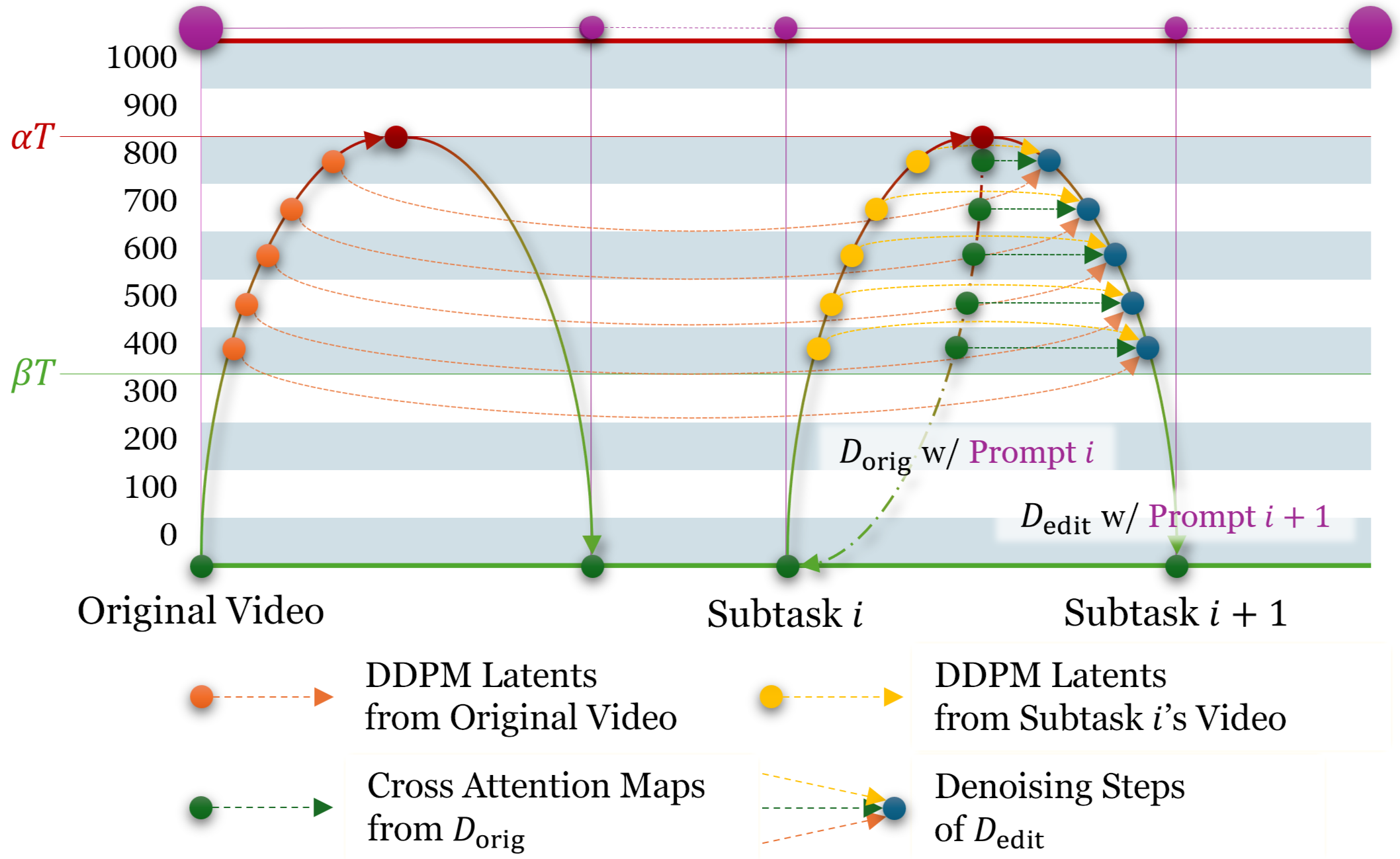}}
\vspace{-3mm}
\caption{\textbf{\themodel preservation control} integrates three key synergistic methods to preserve the original content during editing: (i) control of the initial noise (`$\alpha T$'), (ii) management of noise added at each denoising step (`DDPM Latents'), and (iii) utilization of cross-attention maps between text prompts and video content. Each generation receives guidance on preservation from the previous subtask and the original video for a smooth progression.}
\vspace{-6.5mm}
\label{fig:method:similarity}
\end{figure}

\label{sec:method:similarity}
To preserve the original video content during editing, \themodel{} employs three complementary control mechanisms: (1) controlling the initial noise to maintain low-frequency information; (2) regulating the noise added at each denoising step to preserve semantic details; and (3) utilizing cross-attention maps to ensure alignment between text prompts and video content. These mechanisms work synergistically to maintain the integrity of the original video while enabling effective edits, ensuring successful progression across various editing tasks. A visualization of our preservation control method is shown in Fig.~\ref{fig:method:similarity}.

\noindent\textbf{Basic Formulation.} We utilize a diffusion model with $T$ noise-adding steps to generate videos in formats such as RGB or latent representations, referred to as ``noisy videos'' for generality. The noise-adding steps are denoted as $A_1, A_2, \dots, A_T$, and the denoising steps as $D_T, D_{T-1}, \dots, D_1$. Each denoising step $D_i$ involves a denoising network and a noise scheduler. Let $v_i$ represent the video after the $i$-th noise-adding step, where $v_0$ is the original video and $v_T$ is pure Gaussian noise. Formally, we define $v_i = A_i(v_{i-1})$ and $v_{i-1} = D_i(v_i)$ for $i = 1, 2, \dots, T$.

\noindent\textbf{Initial Noise Control.} To preserve the original video's overall layout during editing, \themodel{} controls the initial noise in the diffusion process. Inspired by SDEdit~\cite{sdedit}, instead of starting generation from pure Gaussian noise with $T$ denoising steps, we limit noise addition to the first $\alpha T$ steps and then perform denoising from this controlled noise with $\alpha T$ denoising steps. This approach maintains low-frequency information, such as the video's structure and layout, while allowing higher-frequency details to be destroyed and regenerated.

\noindent\textbf{Per-Step Noise Control.} The method above inspires us that the noise a diffusion model uses also carries semantic information. Building on this observation, \themodel{} leverages the noise added at each denoising step to preserve the original video content. Specifically, we utilize DDPM inverse~\cite{ddpminv} to extract DDPM latents $n_1, n_2, \dots, n_{\alpha T}$ from the original video during the initial $\alpha T$ noise-adding steps, which are the noises to be added at each step in a DDPM denoising procedure. These latents encapsulate rich semantic details essential for maintaining the video's integrity. By applying them in the corresponding denoising steps $D_{\alpha T}, \dots, D_{\beta T}$, we ensure that semantic information is preserved without excessively constraining high-frequency details, allowing for effective and smooth edits.

However, the DDPM scheduler is inefficient for practical applications. To address this, we explore the intrinsic properties of DDPM inverse, which involves constructing noisy videos and solving for the precise noise required to denoise each step. By defining the denoising function as $D_i(v_i \mid n_i) \xlongequal{\mathrm{Def}} D_i(v_i) + n_i$ for schedulers that do not require random noise $n_i$, we adapt our preservation control to more advanced and efficient schedulers like DDIM~\cite{ddim} and DPMSolver++~\cite{dpmsolver,dpmsolverpp}. This novel adaptation allows \themodel{} to benefit from the semantic preservation capabilities of DDPM inverse, while leveraging the high efficiency of advanced denoising schedulers

\noindent\textbf{Cross-Attention Maps for Generation Control.} To further ensure the preservation of the original video's semantic content, \themodel{} manipulates cross-attention maps in the noise predictor model to align the edited scene's generation process with the original scene. Inspired by attention map replacement strategies of prompt-to-prompt \cite{p2p}, our approach involves simultaneously performing two generations: $D_{\mathrm{orig}}$, which generates the original video using the prompt of the original video; and $D_{\mathrm{edit}}$, which generates the edited video using the prompt of the edited video. By controlling both generations concurrently, we can maintain a consistent semantic alignment between the original and edited content.

To address the high memory and computational cost of naively storing and replacing attention maps, we adopt a fast and memory-efficient approach inspired by Flash Attention~\cite{flashattn}. This technique allows us to compute the attention outputs for both $D_{\mathrm{orig}}$ and $D_{\mathrm{edit}}$ simultaneously during each cross-attention computation through the denoising process, enabling \emph{real-time} replacement of cross-attention maps without the need to store them separately. By further combining the techniques in Flash Attention, we even eliminate the explicit construction of the cross-attention map and directly compute the final outputs. As a result, our method reduces the memory complexity from quadratic to constant, and achieves a fourfold speedup compared to explicit attention map storage and replacement. This efficient implementation ensures that \themodel{} can seamlessly integrate cross-attention map controls within the progressive editing framework, maintaining high-quality and semantically consistent video edits without compromising performance.

\noindent\textbf{Synergy Between Control Mechanisms.} While the three preservation control mechanisms -- initial noise control, intermediate noise control, and cross-attention map control -- can be applied independently, their combination significantly enhances editing performance. By defining an interval $[\beta T, \alpha T]$ during the denoising steps, our framework enables all three controls to operate synergistically \emph{within} this interval. \emph{Outside} of this interval, low-frequency details are preserved, and high-frequency textures are refined, ensuring both content preservation and high-quality edits.

The video editing procedure is shown in Fig. \ref{fig:method:similarity}: First, we perform the initial $\alpha T$ noise-adding steps to obtain the noisy video $v_{\alpha T}$. Then, we simultaneously generate two versions of the video, $D_{\mathrm{orig}}$ using the original prompt and $D_{\mathrm{edit}}$ using the editing prompt. During denoising steps between $\beta T$ and $\alpha T$, the DDPM latents and cross-attention maps from $D_{\mathrm{orig}}$ guide the generation of $D_{\mathrm{edit}}$. For denoising steps before $\beta T$, the model freely refines the video textures without additional guidance.

\subsection{Progression-Based Editing Process}
\label{sec:method:progress}

Different editing tasks may require different levels of preservation control. A mild and easy editing task can succeed with either a lower or higher level of preservation control, but a more challenging editing task that significantly changes the appearance may fail when the preservation control is too strict. To address the varying preservation control requirements across different editing tasks, \themodel{} adopts a progression-based strategy, which decomposes a complex editing task into a sequence of simpler subtasks. As each decomposed subtask is mild and easy to achieve the trade-off between original content preservation and editing task fulfillment, this decomposition allows us to apply a \emph{consistent} preservation control strategy across all subtasks without the need for task-specific adjustments.

As shown in Figs. \ref{fig:method:pipeline} and \ref{fig:method:similarity}, during the progression, for each subtask, \themodel{} simultaneously performs \emph{two} generations: $D_{\mathrm{orig}}$, which regenerates the current subtask video using the original prompt, and $D_{\mathrm{edit}}$, which generates the next subtask video using the editing prompt. The generation of $D_{\mathrm{edit}}$ is guided by the cross-attention maps and DDPM latents extracted from both $D_{\mathrm{orig}}$ and the original video generation (``$D_{\mathrm{orig video}}$'') with a mixture coefficient.

By progressively completing each subtask with this dual-guided generation, \themodel{} maintains high-quality and semantically consistent edits across various scenarios. 
This synergistic approach effectively balances the preservation of original content with the fulfillment of editing instructions, ensuring smooth and successful progression from one subtask to the next without the complexity of designing different levels of control mechanisms.

\subsection{Efficient and Stable 3D Scene Editing}
\label{sec:method:3d}

Beyond its native video editing capabilities, \themodel{} seamlessly extends to 3D scene editing by incorporating a straightforward \emph{render-edit-reconstruct (RER)} process: Render a video of the original scene along a fixed camera trajectory, perform video editing using \themodel, and then reconstruct and re-render the scene from the edited video. 

To ensure 3D consistency, we modify the progressive editing framework so that after obtaining the edited video for each subtask, we could reconstruct it to 3D and re-render it back to video for the next subtask. This modification leverages both the temporal smoothness of the rendered video and the 3D consistency of reconstruction, ensuring strong 3D consistency in the edited video. Unlike previous 3D editing methods that require iterative dataset updates and additional training, our approach remains stable and efficient, enabling high-quality edits with minimal diffusion generations. Furthermore, the temporal consistency of our edited videos allows for significant geometric changes, such as object insertion, which were previously challenging due to inconsistent per-view editing results.

\section{Experiment}
\label{sec:expr}

\subsection{Experimental Settings}

\begin{figure*}[t!]
\centering
\centerline{\includegraphics[width=1\linewidth]{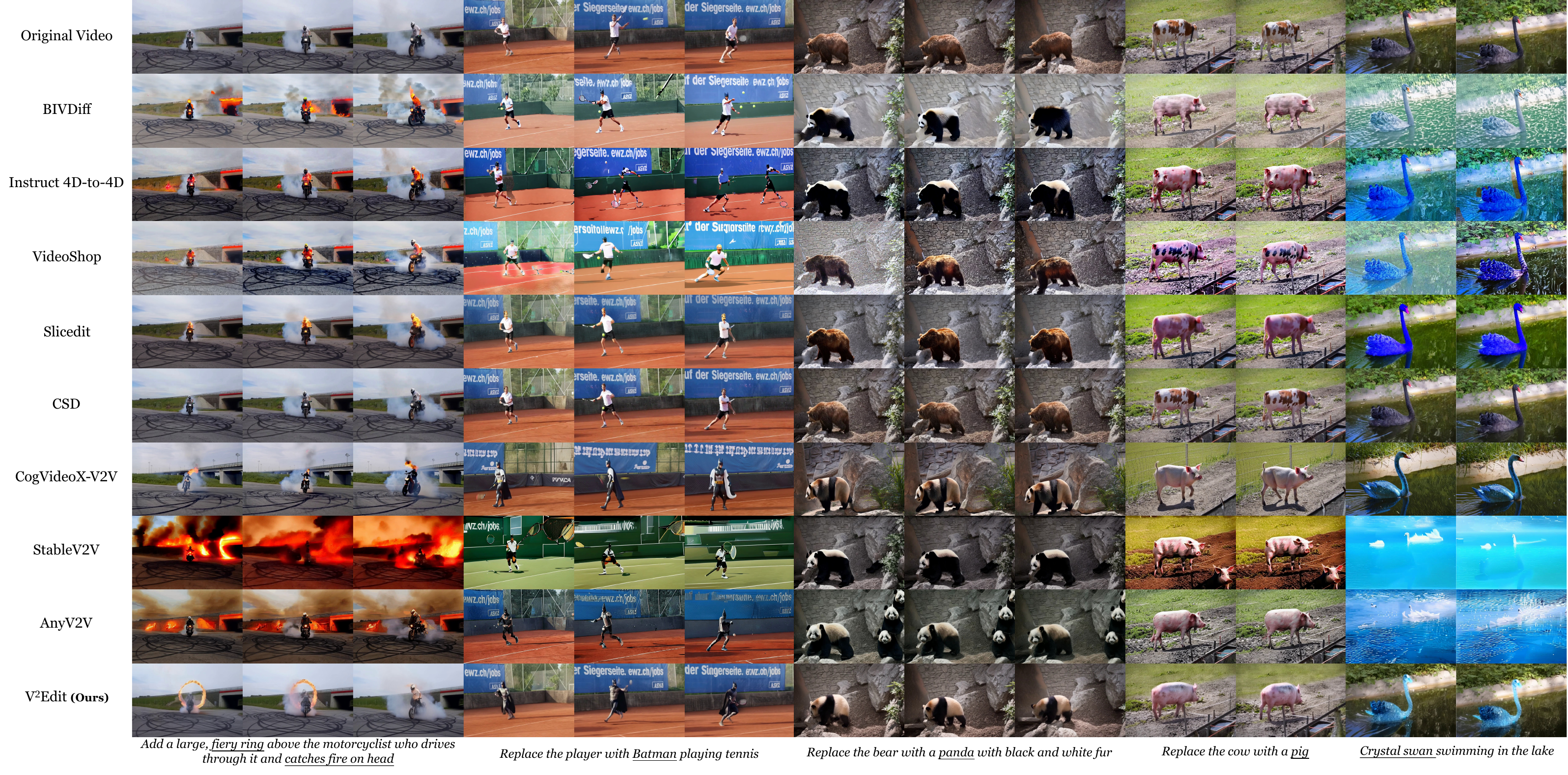}}
\vspace{-3mm}
\caption{Our \themodel achieves successful editing results in various video editing tasks with superior overall appearance, while well preserving the original contents. The baselines either generate results with strange appearance and artifacts, or fail to preserve the areas unrelated to the editing. Notably, CogVideoX-V2V \cite{cogvideox}, an official video-to-video editing model of CogVideoX, generates good-looking results but is unable to preserve the original contents, showing that the key of our \themodel lies in our novel progression framework and preservation control mechanism, instead of the strong underlying CogVideoX model. \textbf{More results are on \href{https://immortalco.github.io/V2Edit/}{our project website}.}}
\vspace{-3mm}

\label{fig:expr:vid}
\end{figure*}

\begin{figure*}[t!]
\centering
\centerline{\includegraphics[width=1\linewidth]{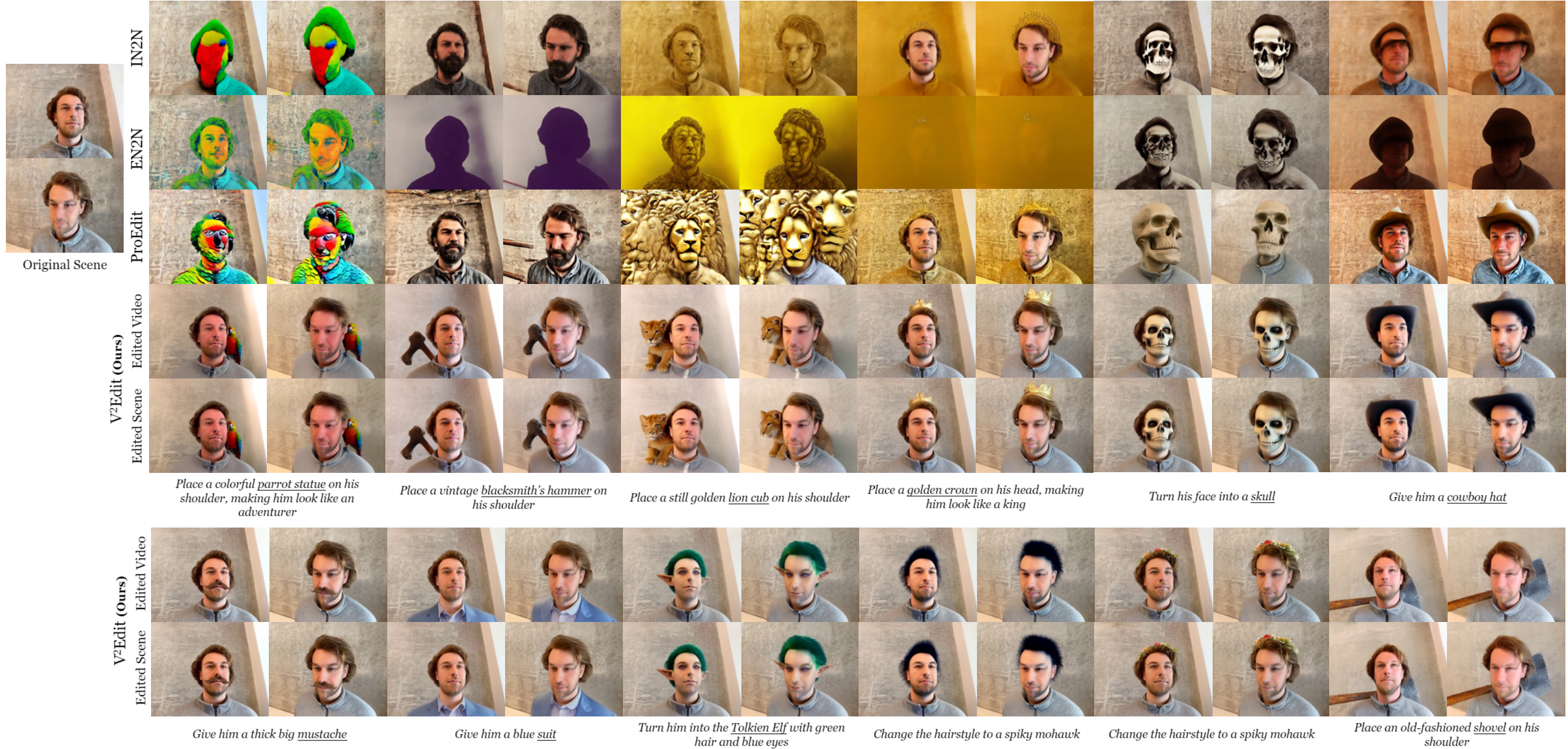}}
\vspace{-3mm}
\caption{Our \themodel achieves high-quality editing results in various challenging 3D scene editing tasks in the Face scene of the IN2N \cite{in2n} dataset, with clear texture and geometry structure, bright color, and superior original content preservation. Notably, our \themodel successfully performs editing operations with significant geometric changes like object insertion. On the contrary, the baselines either fail to perform the editing or do not preserve the contents in the original scene, \eg., the background color, the appearance of the person, \etc.}

\vspace{-6mm}
\label{fig:expr:face}
\end{figure*}

\begin{figure*}[t!]
\centering
\centerline{\includegraphics[width=1\linewidth]{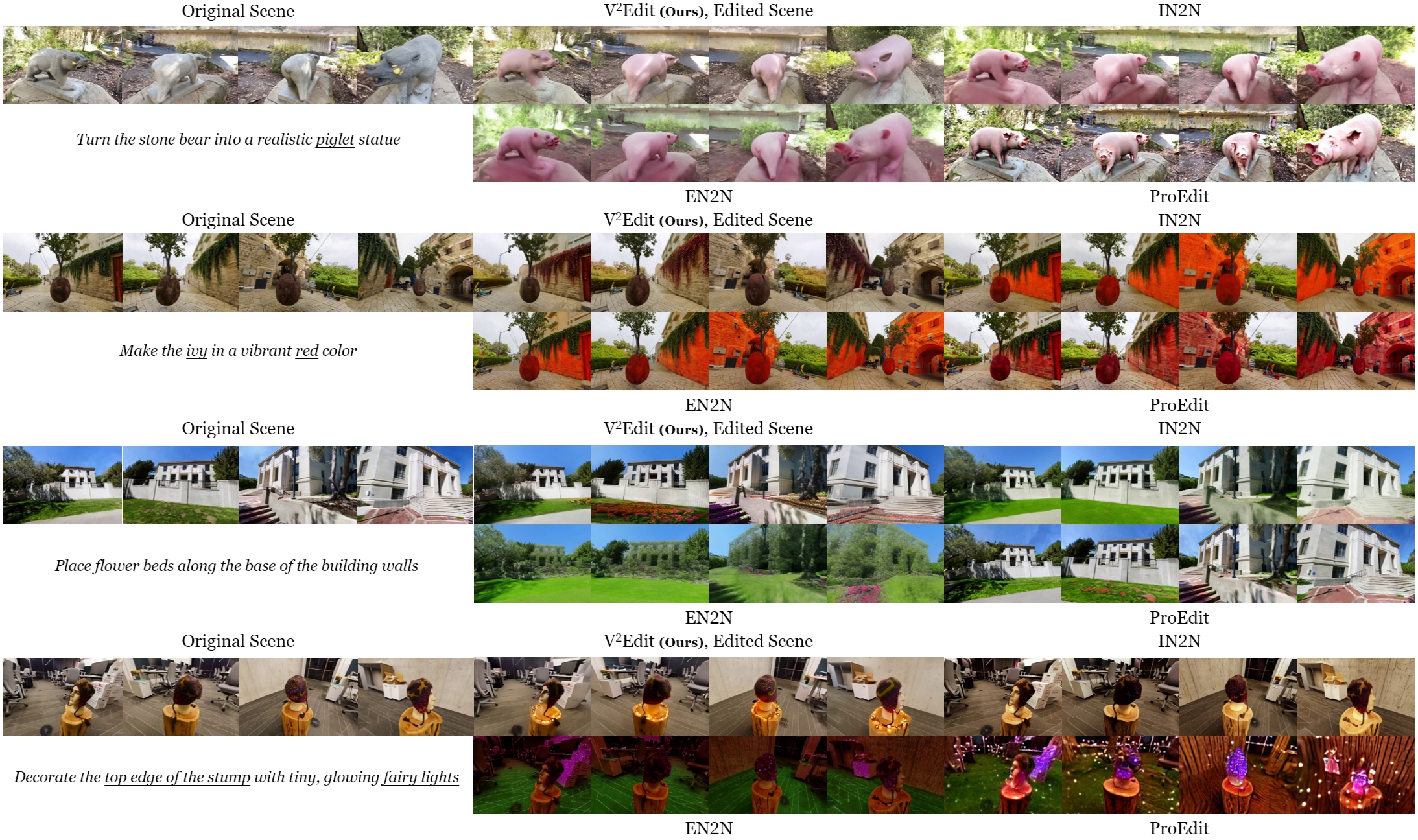}}
\vspace{-3mm}
\caption{Our \themodel achieves high-quality editing results across various indoor and outdoor scenes, while consistently achieving both editing task fulfillment and original content preservation for all the tasks. The baselines either fail to perform the editing or edit many unrelated areas without satisfactory preservation. \textbf{More results are on \href{https://immortalco.github.io/V2Edit/}{our project website}.}}
\vspace{-6.5mm}
\label{fig:expr:3d}
\end{figure*}

\begin{table}[t!]
\centering
\scalebox{0.65}{\begin{tabular}{l|cccc}
 \hline\hline
 Method & CTIDS $\uparrow$ & CDC $\uparrow$ & GPT Score $\uparrow$ & User Study $\uparrow$ \\
 \hline 
 \multicolumn{5}{c}{Video Editing}\\
 \hline 
 BIVDiff \cite{bivdiff} & 0.0755 & 0.1007 & 73.95 & 2.43 \\
 Instruct 4D-to-4D \cite{i4d24d} & 0.0502 & 0.0069 & 75.09 & 2.33 \\
 VideoShop \cite{videoshop} & 0.0489 & 0.0967 & 60.00& 2.29 \\
 Slicedit \cite{slicedit} & 0.2867 & 0.1332 & 74.04 & 2.17 \\
 CSD \cite{csd} & 0.1708 & 0.0534 & 48.48 & 2.14 \\

 \themodel \textbf{(Ours)} & \textbf{0.3098} & \textbf{0.1388} & \textbf{84.50} & \textbf{3.83} \\
\hline 
 \multicolumn{5}{c}{3D Scene Editing}\\
 \hline 
Instruct-NeRF2NeRF \cite{in2n} & 0.0542 & 0.1468 & 55.81 & 2.30 \\
Efficient-NeRF2NeRF \cite{en2n} & 0.0367 & 0.1389 & 56.48 & 2.12 \\
ProEdit \cite{proedit} & 0.1048 & 0.1150 & 48.65 & 2.42 \\
\themodel \textbf{(Ours)}, Edited Scene & \textbf{0.2081} & \textbf{0.1716} & \textbf{88.97} & \textbf{3.90} \\

 \hline\hline
\end{tabular}}
\vspace{-3mm}
\caption{Quantitative evaluation shows that our \themodel consistently outperforms all the baselines under all metrics in both video and 3D scene editing tasks.}

\vspace{-8mm}
\label{tab:exp:quan}
\end{table}

\begin{figure}[t!]
\centering
\centerline{\includegraphics[width=1\linewidth]{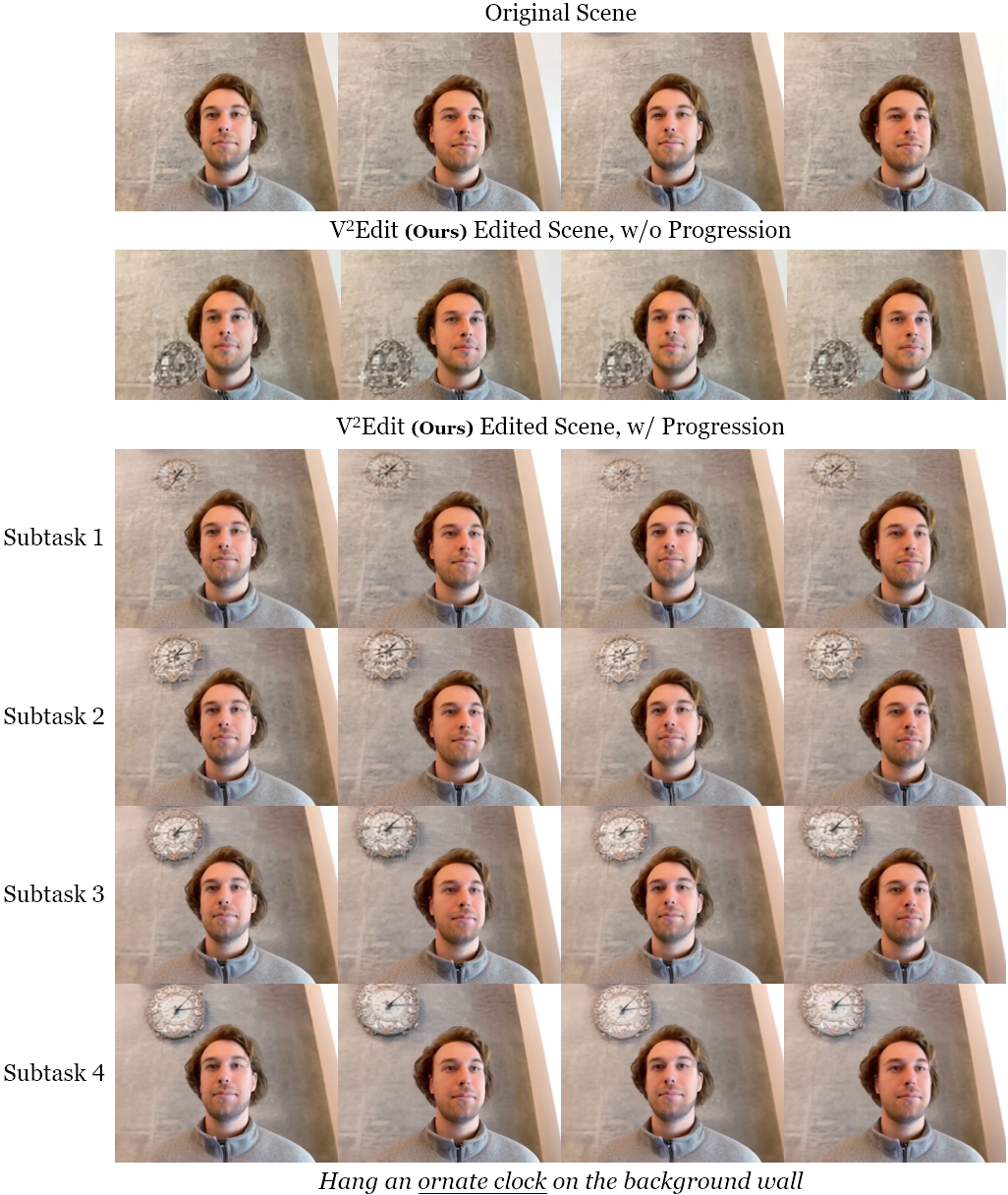}}
\vspace{-3mm}
\caption{Ablation study on our progressive framework shows that progression is crucial to obtain high-quality editing results. Notably, the progressive editing process demonstrates a gradual procedure of editing to construct the clock on the wall, ending up with a clock with even 3D-consistent clock hands. }
\vspace{-6.5mm}
\label{fig:expr:abla:prog}
\end{figure}

\noindent\textbf{\themodel Settings.} We use CogVideoX-5b \cite{cogvideox} as the underlying video diffusion model, which is a text-to-video model based of a diffusion transformer (DiT), and supports SORA-like \cite{sora} long descriptions as input prompts. We use GPT-4o \cite{gpt4o} as the LVLM to generate the prompts for underlying CogVideoX. For the subtask decomposition in our progressive framework, we allow at most six (6) subtasks for each editing task. For 3D scene editing tasks, our \themodel is independent of the specific scene representation. Therefore, we choose either SplactFacto or NeRFacto from NeRFStudio \cite{nerfstudio} as our scene representation.

\noindent\textbf{Video Editing Tasks.} Consistent with previous work \cite{bivdiff}, we use the videos from DAVIS dataset \cite{davis,daviscompetition} as the source videos. The editing tasks for evaluation are suggested by GPT-4o, given the original video as input.

\noindent\textbf{Video Editing Baselines.} We compare our \themodel with video editing baselines, which can be roughly divided into two categories: (1) Image-based methods which rely on an underlying image generative model, including Slicedit \cite{slicedit}, and Instruct 4D-to-4D \cite{i4d24d} for monocular scenes; and (2) Video-based methods with utilizes an underlying video generative model, including CogVideoX-V2V \cite{cogvideox}, VideoShop \cite{videoshop}, StableV2V \cite{stablev2v}, AnyV2V \cite{anyv2v}, BIVDiff \cite{bivdiff} with per-frame editing and overall refinement, and CSD \cite{csd}. Some image-based methods require the edited first frame as a guidance, and we consistently apply Instruct-Pix2Pix \cite{ip2p} to generate this frame.

\noindent\textbf{3D Scene Editing Tasks.} Consistent with previous scene editing methods \cite{in2n,proedit,igs2gs,gausseditor}, we mainly use the scenes in Instruct-NeRF2NeRF (IN2N) \cite{in2n} dataset for comparison evaluations. We also use several outdoor scenes from NeRFStudio \cite{nerfstudio} to serve as more challenging tasks. For the camera trajectory of the scene, we either use the existing trajectories (for IN2N dataset with officially provided trajectories) or manually draw one (for other scenes).

\noindent\textbf{3D Scene Editing Baselines.} We compare our \themodel with state-of-the-art traditional image-based 3D scene editing methods, including Instruct-NeRF2NeRF (IN2N) \cite{in2n}, Efficient-NeRF2NeRF \cite{en2n}, and ProEdit \cite{proedit}. In the \textbf{supplementary}, we also compare another type of baseline: applying the RER strategy (Sec. \ref{sec:method:3d}) with the video editing baselines mentioned above.

\noindent\textbf{\themodel Variants for Ablation Study.} In the main paper, we provide the ablation study with the following key \themodel variants: 
(1) CogVideoX-V2V, which also utilizes CogVideoX \cite{cogvideox} as the underlying video diffusion model;
(2) No Progression (NP), which only applies our original preservation control to the editing without progression.
Due to limited space, we provide more ablation study results in the \textbf{supplementary} with more variants.

\noindent\textbf{Metrics.} The evaluation of video editing tasks contains many aspects, including the overall visual quality, the original video preservation, and the editing task fulfillment. It is challenging to evaluate them using traditional methods. Therefore, consistent with \cite{proedit},  we use GPT-4o \cite{gpt4o} for this evaluation, which can be regarded as a Monte Carlo simulation of the VQAScore \cite{vqascore}. We provide GPT with the requirements of each aspect, the editing instruction, and the original and edited videos frame-by-frame, and then ask GPT to provide a score from 1 to 100 for each aspect. To compare multiple videos between ours and different baselines, we provide all these videos simultaneously to GPT, and ask GPT to score them all together to enforce a consistent scoring rule. \updated{To avoid randomness, we use the average of 20 independent evaluations as the final results. } Leveraging the vision-language reasoning capability of GPT, this metric can quantify different aspects of the edited video. We also provide user study and the CLIP \cite{clip}-based scores from \cite{in2n}: CLIP Text-Image Direction Similarity (CTIDS), and CLIP Direction Consistency (CDC). 

\subsection{Experimental Results}
\noindent\textbf{Video Editing.} The visualization results of video editing on DAVIS \cite{davis} dataset are in Fig. \ref{fig:expr:vid}, while more results are on \href{https://immortalco.github.io/V2Edit/}{our project website}. Our \themodel consistently edits successfully and produce high-fidelity results in various challenging tasks, \eg, adding a fiery ring for the motorcyclist to drive through, and turn a fast-moving person into a Batman; while successfully preserving the unrelated part, \eg, the wall and layout of the tennis court and the tennis player's motion in ``Batman'' task, the objects in the farm in ``pig'' task, and the river in the ``swan'' task. On the contrary, each baseline either fails to perform the editing or is unable to preserve the unrelated part from the original scene -- especially the original pose and motion. Notably, the baseline CogVideoX-V2V is an official method that applies SDEdit \cite{sdedit} on CogVideoX, which can be regarded as a variant of ours. This baseline produces videos with good appearance, but fails to preserve most of the information from the original scene. This validates the cruciality of our preservation control method. This shows that it is not the strong capability of the underlying CogVideoX we use, but our novel original preservation and progression pipeline that leads to our high-quality editing results.

\noindent\textbf{3D Scene Editing.} The results of 3D scene editing are shown in Figs. \ref{fig:expr:face} and \ref{fig:expr:3d}, and more results are on \href{https://immortalco.github.io/V2Edit/}{our project website}. As shown in Fig. \ref{fig:expr:face}, our \themodel succeeds in challenging editing tasks that contain significant geometric change, with clear appearance and reasonable geometry structure, especially in the ``lion cub'' editing. \eg, object insertion, while all the baseline fails to perform most these tasks -- either unable to fulfill the editing requirement or completely changes the appearance of the original scene, or both. Despite of the face-forward scene, our \themodel also performs well in the indoor or outdoor scenes in Fig. \ref{fig:expr:3d} in diversified editing instructions, with both great fulfillment of editing instruction and preservation of the original scene. Notably, with our self-implemented flash-attention-based \cite{flashattn} acceleration in Sec. \ref{sec:method:similarity}, editing a 72-frame video only takes 10 minutes for each subtask in the progressive framework. Therefore, one editing task with at most six progression subtasks only takes roughly one to two hours to perform, achieving comparable efficiency as simple baselines \cite{in2n,en2n} but producing significantly superior results.

\noindent\textbf{Quantitative Evaluations.}  We perform quantitative evaluations on several representative editing tasks, with results presented in Tab.~\ref{tab:exp:quan}, including a user study involving 43 participants conducted to assess subjective quality. Our \themodel{} consistently outperforms all baseline methods across all metrics in both video and 3D scene editing. Specifically, \themodel{} successfully balances original content preservation, as measured by the `CDC' metric, which quantifies adjacent-frame similarity between the original and edited scenes; and editing task fulfillment, as demonstrated by GPT-based evaluations and user study results. These findings establish \themodel{} as a state-of-the-art framework in both video and 3D scene editing domains.

\noindent\textbf{Ablation Study.} As illustrated in Fig.~\ref{fig:expr:vid}, the baseline CogVideoX-V2V generates high-quality videos across various editing tasks but consistently fails to preserve unrelated content from the original video. This baseline effectively represents a variant of our \themodel{} with only initial noise control. These results indicate that leveraging a powerful video diffusion model alone is insufficient for high-quality editing without an effective content preservation mechanism, underscoring the necessity of our preservation control strategy. Additionally, as shown in Fig.~\ref{fig:expr:abla:prog}, directly applying our content preservation mechanism without the progression framework results in failures in complex tasks, such as adding a clock. In contrast, when incorporating the progression-based editing strategy, \themodel{} successfully constructs and refines the clock, achieving high-quality results. Notably, the clock hands remain consistent across all views, demonstrating excellent 3D consistency. These experiments validate that both our content preservation mechanism and progression framework are essential, which not only ensure content preservation but also achieve editing task fulfillment. Further ablation study results are provided in the \textbf{supplementary}.

\section{Conclusion}
\label{sec:conclu}

In this paper, we introduce \themodel{}, a novel and versatile framework for instruction-guided video and 3D scene editing. Our approach effectively balances the preservation of original content with the fulfillment of editing instructions by progressively decomposing a complex task into simpler subtasks, managed by a unified preservation control mechanism. For video editing, \themodel{} excels in handling challenging scenarios involving fast-moving camera trajectory, complex motions, and significant temporal variations, ensuring smooth and consistent edits. For 3D scene editing, our framework supports challenging editing tasks with substantial geometric changes, while maintaining high 3D consistency and sufficiently preserving the original scene content. Extensive experiments demonstrate that \themodel{} achieves state-of-the-art performance in both video and 3D scene editing. We hope that \themodel{} paves the way for future advancements in video and 3D scene editing using video diffusion models. \hspace{-1000mm}
 
{
    \small
    \bibliographystyle{ieeenat_fullname}
    \bibliography{main}
}

\clearpage
\setcounter{page}{1}
\maketitlesupplementary

\appendix
\renewcommand{\thefigure}{\Alph{figure}}
\renewcommand{\thetable}{\Alph{table}}

This document contains additional analysis and extra experiments.



\section{Additional Experimental Results}

\subsection{Additional Visualizations}

\begin{figure*}[t!]
\centering
\centerline{\includegraphics[width=1\linewidth]{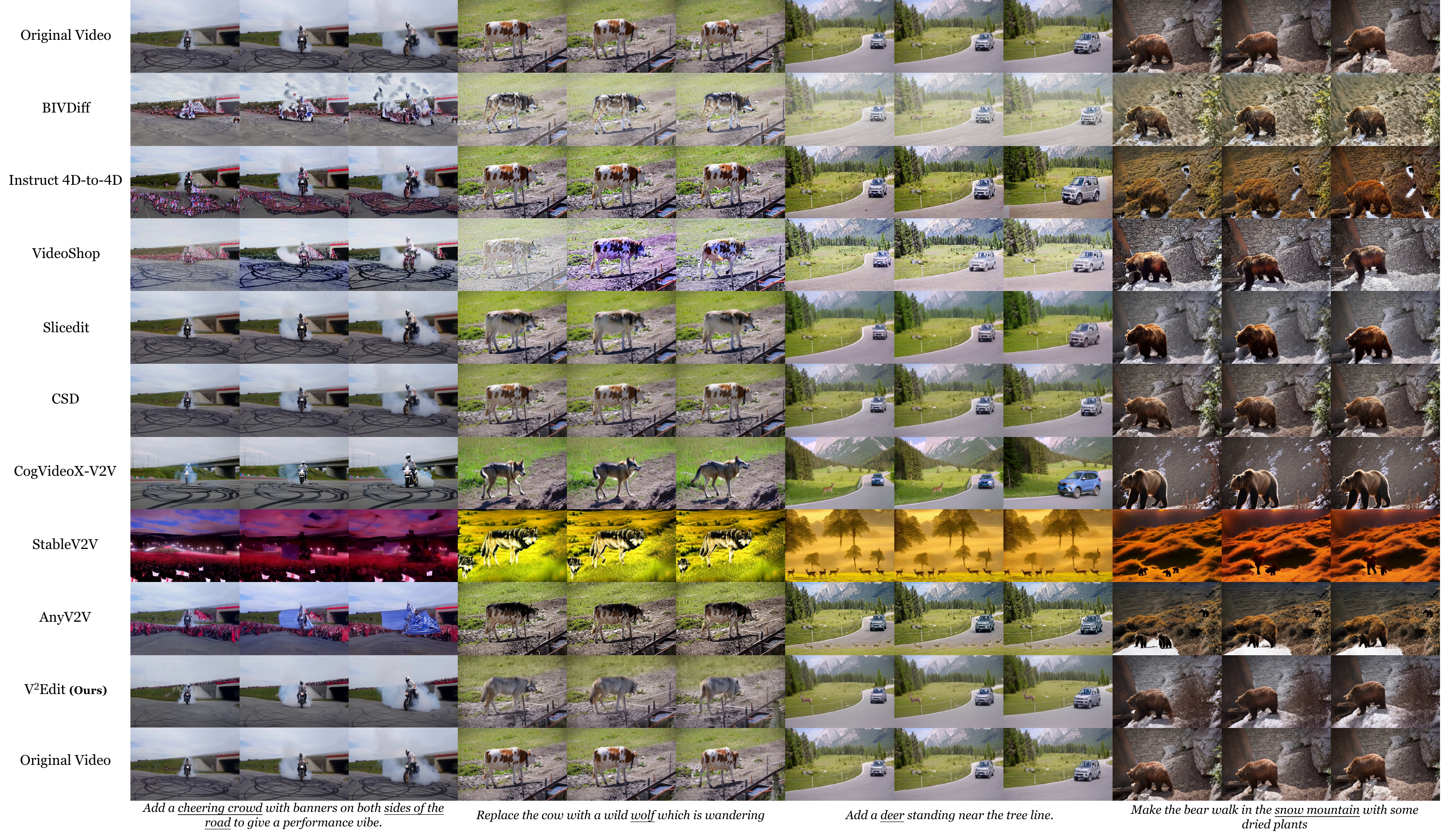}}
\caption{\textbf{In additional video editing tasks, our \themodel framework} consistently outperforms baselines in both editing task fulfillment and original content preservation.}
\label{fig:suppl:morevid}
\end{figure*}

\begin{figure*}[t!]
\centering
\centerline{\includegraphics[width=1\linewidth]{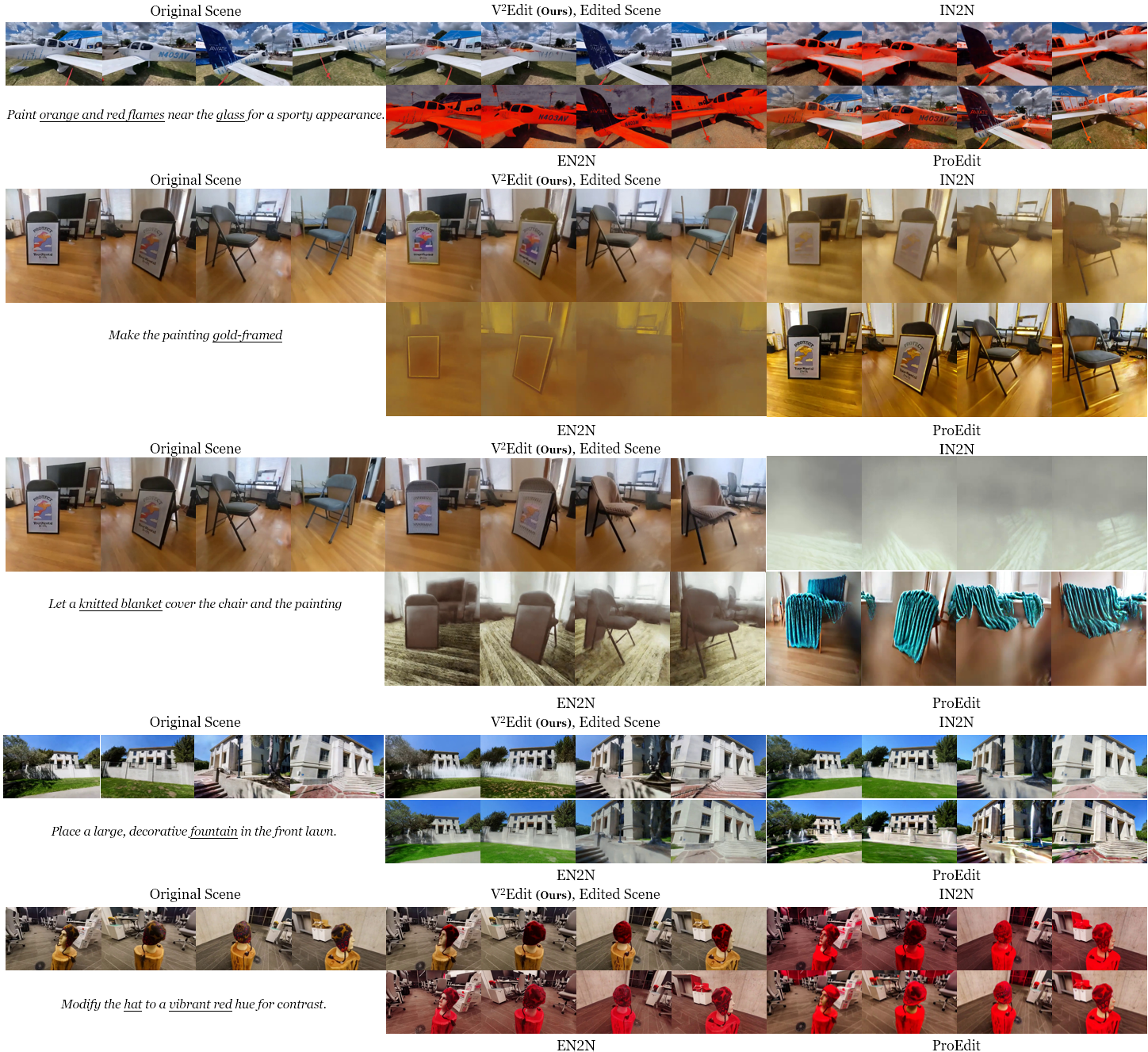}}
\caption{\textbf{In additional 3D scene editing tasks, our \themodel framework} consistently outperforms baselines in both editing task fulfillment and original content preservation.}
\label{fig:suppl:more3d}
\end{figure*}

We show additional visualization results for video editing in Fig. \ref{fig:suppl:morevid}, and for 3D scene editing in Fig. \ref{fig:suppl:more3d}. Our \themodel framework consistently outperforms baselines in both editing task fulfillment and original content preservation in all these tasks.

\subsection{Video Editing Baselines for 3D Scene Editing}

\begin{table*}[t!]
\centering
\scalebox{1}{\begin{tabular}{l|ccc}
 \hline\hline
 Method & CTIDS $\uparrow$ & CDC $\uparrow$ & GPT Score $\uparrow$ \\
 \hline 
 \multicolumn{4}{c}{Edited Video (Intermediate Results of 3D Scene Editing)}\\
 \hline 
 BIVDiff \cite{bivdiff} & 0.1656       & 0.0813     & 51.00 \\
 Instruct 4D-to-4D \cite{i4d24d} & 0.0224       & 0.1723     & 29.60 \\
 VideoShop \cite{videoshop} & 0.0068       & -0.0389    & 21.60 \\
 Slicedit \cite{slicedit} & -0.0064      & 0.1817     & 31.00 \\
 CSD \cite{csd} & -0.0035      & 0.1930     & 30.60 \\
 CogVideoX-V2V \cite{cogvideox} & 0.2125       & 0.0554     & 67.80 \\

 \themodel \textbf{(Ours)} & \textbf{0.2796} & \textbf{0.1934} & \textbf{90.20} \\
\hline 
 \multicolumn{4}{c}{Rendered Edited Scenes}\\
 \hline 
 BIVDiff \cite{bivdiff} &0.0901       &0.0691     &44.00 \\
 Instruct 4D-to-4D \cite{i4d24d}&-0.0117      &0.1507     &25.20 \\
 VideoShop \cite{videoshop} &0.0045       &0.0271     &14.20 \\
 Slicedit \cite{slicedit} &-0.0061      &0.1728     &24.00 \\
 CSD \cite{csd} &0.0035       &0.1950     &23.60 \\
 CogVideoX-V2V \cite{cogvideox} &0.2042       &0.0266     &52.80 \\

 \themodel \textbf{(Ours)} & \textbf{0.2817} & \textbf{0.2012} & \textbf{90.60} \\

 \hline\hline
\end{tabular}}

\caption{Quantitative evaluation shows that our \themodel significantly outperforms all the baselines that use video editing methods for 3D scene editing. This validates that our \themodel is uniquely capable of performing 3D scene editing among video-based methods.}

\label{tab:supp:3dtype2}
\end{table*}

The qualitative results of video editing baselines for 3D scene editing are on our project website, and the quantitative results are in Tab. \ref{tab:supp:3dtype2}. Our \themodel significantly outperforms all these methods, especially in the GPT score, which evaluates the overall quality. Also, as shown in the videos on our project website, even when the videos edited by these baselines have reasonable appearance, the 3D reconstruction is still low-quality and contains lots of artifacts. This shows that our \themodel is the only model that has sufficient editing capability and preservation control ability to edit 3D scenes.

\subsection{Ablation Study}

\begin{figure*}[t!]
\centering
\centerline{\includegraphics[width=1\linewidth]{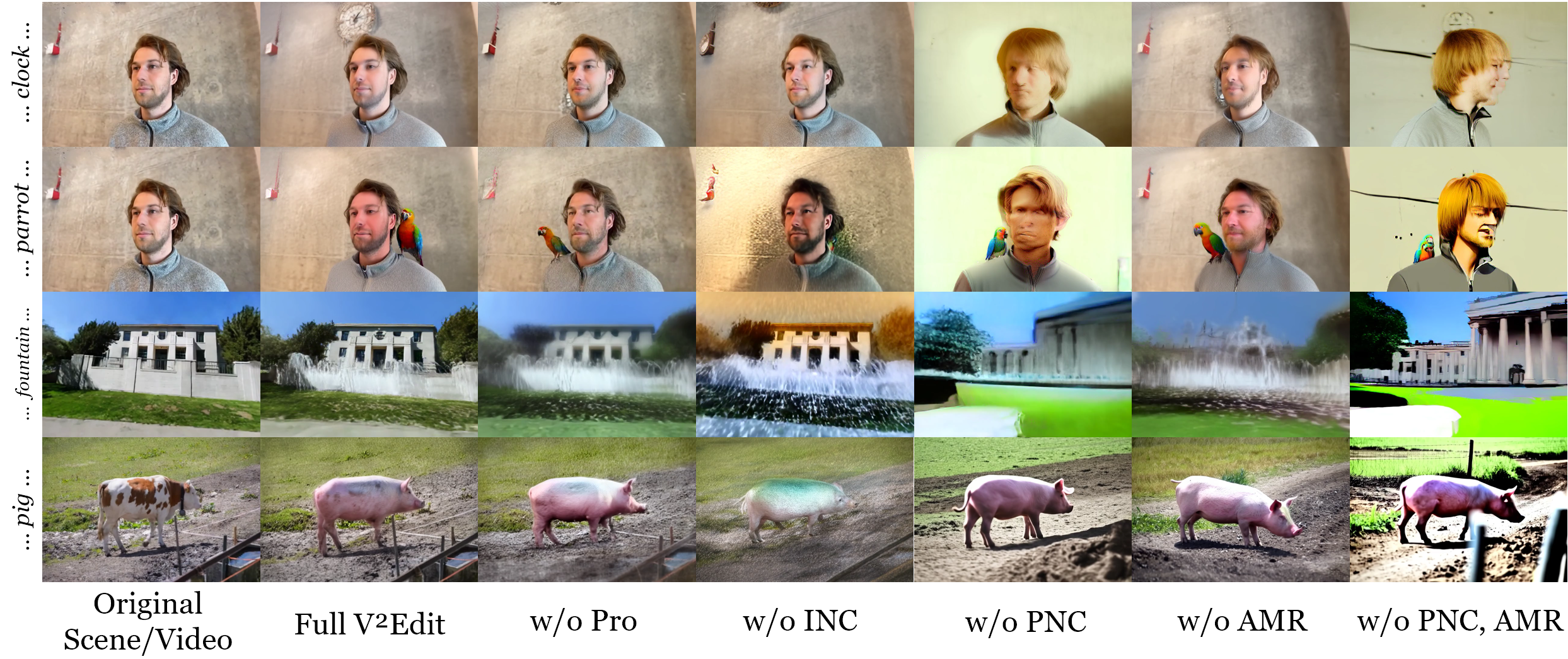}}
\caption{Ablation studies show that all our strategies are crucial to successful and high-quality editing results. Removing any of them results in significant degradation. The video visualizations are on our project website. }
\label{fig:suppl:abla}
\end{figure*}

\begin{table*}[t!]

\centering
\scalebox{1}{\begin{tabular}{l|ccc}
 \hline\hline
 Method & CTIDS $\uparrow$ & CDC $\uparrow$ & GPT Score $\uparrow$ \\
 \hline 
 \multicolumn{4}{c}{Video Editing}\\
 \hline 
 w/o Progression (Pro) &0.0611       &0.2414     &65.23\\
 w/o Initial Noise Control (INC) &0.0763       &0.1922     &64.80\\
 w/o Per-Step Noise Control (PNC) &0.0949  &0.0291     &55.83\\
 w/o Attention-Map Replacement (AMR) &0.0898       &0.1079     &51.04\\
 w/o PNC, AMR &0.0674       &0.0394     &40.21\\
 Full \themodel \textbf{(Ours)} & \textbf{0.1029} & \textbf{0.2933} & \textbf{76.60} \\
\hline 
 \multicolumn{4}{c}{3D Scene Editing}\\
 \hline 
 w/o Progression (Pro) &0.0571       &0.2744     &59.17\\
 w/o Initial Noise Control (INC) &0.0655       &0.2462     &52.50\\
 w/o Per-Step Noise Control (PNC) &0.0789       &0.0088     &34.17\\
 w/o Attention-Map Replacement (AMR) &0.0829       &0.1109     &53.33\\
 w/o PNC, AMR &0.0631       &0.0370     &41.67\\
 Full \themodel \textbf{(Ours)} & \textbf{0.0954} & \textbf{0.3658} & \textbf{77.50} \\

 \hline\hline
\end{tabular}}

\caption{Ablation study shows that all our strategies are crucial to our final performance.}

\label{tab:supp:abla}
\end{table*}

We conduct our ablation study on each of our core strategies: Progression (`Pro'), initial noise control (`INC'), per-step noise control (`PNC'), and attention-map replacement (`AMR'). 

The ablation study results are in Fig. \ref{fig:suppl:abla} and on our project website. The quantitative measurements are in Tab. \ref{tab:supp:abla}. We observe that all these core strategies are crucial to our final results. More specifically:

\begin{itemize}
    \item Progression is crucial to the success and clear appearance of the final results. Without progression, some geometry editing tasks for 3D scenes may even fail.
    \item Per-step noise control (PNC) is the most important control in our original preservation framework. Without PNC, the edited video will be significantly different from the original view, which implies a failure in original preservation.
    \item Initial noise control is crucial to the preservation of the overall color of the edited video.
    \item Attention-map replacement is crucial for the preservation of the shape and appearance of unrelated objects.
    \item The variant ``w/o PNC, AMR'' is equivalent to CogVideoX-V2V \cite{cogvideox} with progression. The failure of this variant shows that CogVideoX-V2V cannot succeed in editing tasks even with progression, further validating that the high-quality results of our \themodel are not from the powerful underlying CogVideoX but our novel strategies.
\end{itemize}

\section{Implementation Details}

\subsection{Settings and Hyperparameters}

\paragraph{GPUs.} All of our experiments are run on NVIDIA A6000, A100, and H100 GPUs. Each instance of editing task only needs one GPU.

\paragraph{Underlying Video Diffusion Model.} We use CogVideoX-5b \cite{cogvideox} as our underlying diffusion model, and fetch the pre-trained weights from HuggingFace. The model is run in pure Brain Float16 (``bfloat16'') data type. We consistently set the classifier-free guidance scale \cite{diffusion} as 7.0, and use CogVideoX's default \texttt{CogVideoXDPMScheduler}.

\paragraph{Hyperparameters.} We choose the hyperparameters for our original preservation control method as follows. We set $\alpha = 0.9, \beta=0.5$. As the $T$ (number of denoising steps) of CogVideoX is 1000, this means that we first perform the first $\alpha T=900$ noise addition steps and then start denoising at this generation, and apply the guidance from DDPM inverse and the cross-attention map replacement until the $\beta T=500$-th denoising step.

\subsection{Flash Attention-Based Optimization for Attention Map Replacement}

In dot-product attention $$\mathrm{Attn}(Q,K,V) = \mathrm{SoftMax}\!\left(QK^\top / \sqrt{d}\right) V,$$
where $Q$ is $n\times d$, $K$ is $m\times d$, and $V$ is $m\times d'$, the most expensive operation is to explicit construct $M=\left(QK^\top / \sqrt{d}\right)$, a.k.a. ``the attention map'', which is $n\times m$. However, the SoftMax operation does not allow us to directly compute the output. The key insight of flash attention \cite{flashattn} optimization is as follows: consider each row of $\mathrm{SoftMax(M)}$, we have $$\mathrm{SoftMax(M)}_{i,j} = \frac{\exp(M_{i,j})}{\sum_j \exp(M_{i,j})}.$$ Therefore, we can first compute the denominator $d_i = \sum_j \exp(M_{i,j})$ for each row $i$, and then directly calculate $\mathrm{SoftMax(M)}_{i,j} = \exp(M_{i,j}) / d_i$. To avoid the precision issue of $\exp(M_{i,j})$ when $M_{i,j}$ is large, we first compute $m_i = \max_j M_{i,j}$, and then define $d'_i=\sum_j \exp(M_{i,j} - m_i)$ to take the place of $d_i$, so that $$\mathrm{SoftMax(M)}_{i,j} = \exp(M_{i,j} - m_i) / d'_i.$$  At last, the final output $$
\begin{aligned}\mathrm{Attn}(Q,K,V)_i =& (\mathrm{SoftMax(M)} V)_i \\ =& \sum_j \mathrm{SoftMax(M)}_{i,j} v_j  \\ =& \sum_j \exp(M_{i,j} - m_i)  v_j / d'_i  \end{aligned}.$$ All the elements $M_{i,j}, m_i, d'_i$ can be computed at the same time complexity but without explicitly constructing the whole matrix $M$, which significantly saves the memory cost by reducing the additional memory complexity from $O(nm)$ to $O(n)$,  and the time to allocate and access the memory.

When we use attention map replacement, the only difference is that for the computation of $M$, some columns (corresponding to $K,V$, which is the prompt token sequences) need to be replaced from another $M'$. In this case, we perform the cross-attention of both operations in parallel, and replace the attention map \emph{on-the-fly}. We define $$\mathrm{AttnAMR}(Q^{(1)},K^{(1)},V^{(1)},Q^{(2)},K^{(2)},V^{(2)},I^{(1)},I^{(2)})$$ as such function, where $\mathrm{Attn1}=(Q^{(1)},K^{(1)},V^{(1)})$ is the attention operation to be computed as usual, and $\mathrm{Attn2}=(Q^{(2)},K^{(2)},V^{(2)})$ is the attention operation to be computed with attention-map replacement (AMR), $I^{(1)},I^{(2)}$ is the index list that the $I^{(2)}_k$-th column of attention map $M^{(2)}$ of $\mathrm{Attn2}$ should be replaced with the $I^{(1)}_k$-th column of attention map $M^{(1)}$ of $\mathrm{Attn1}$ for each $1\le k \le |I^{(1)}|=|I^{(2)}|$. In the computation, we only need to simply replace the computation of $M^{(2)}$ to fit the rule of attention map replacement, \ie, calculate $M^{(2)}_{i,j}$ with $Q^{(1)}_iK^{(1)\top}_{j'}$ instead of $Q^{(2)}_iK^{(2)\top}_j$ if $j = I^{(2)}_k, j'=I^{(1)}_k$ for some $k$. In this way, we extend the optimization of flash attention to the attention with attention-map replacement.

In our experiments, this optimization could bring a 2 $\sim$ 4 times speed-up.

\subsection{Attention Control for Long and Looping Videos}

Most of the video diffusion models are trained to generate fixed-length videos. For example, CogVideoX is trained to generate 49-frame videos at 8 FPS. However, some videos to be edited are longer than this length, especially in 3D editing tasks. If we speed up the video to fit the frames, the camera movement will be too fast and, therefore, introduce many challenges for the editing operation to be successful. To address this, we propose a way to enable the support of long video in a training-free manner.

One naive way to generate a long video with a diffusion model is just to extend the size of the input noise, which will not encounter out-of-memory issue with our flash attention-based optimization. However, as the model is only trained on 49-frame videos, it gets confused with the long-term temporal positional embeddings, which are unseen in the training dataset and may lead to low-quality videos. In the generation, this happens in the self-attention, where both $Q$ and $K$ correspond to video patch tokens, and the temporal embedding of $Q_i$ and $K_j$ are too far away. Also, as the $K$ is much larger than the trained situation, the attention also aggregates too much elements from $K$ and further lead to blurred results.

Our insight is that, we can constraint the set of video tokens $\{K_j\}$ that can be seen by each video token $Q_i$, so that it will only encounter the patterns of positional embeddings that seen in training, and also not see and aggregate too many $K_j$ to make the video blurred. More specifically, if the $Q_i$ belongs to frame $k$, we allow $Q_i$ to see all the tokens within frames $[k-l/2,k+l/2]$, where $l$ is the length of the pre-trained videos. With this control, each $Q_i$ will only see the tokens within nearby $l$ frames, which are seen patterns in training, and also not aggregating too much $\{K_j\}$s -- which will be no more than $l$ frames, \ie the number of $\{K_j\}$s seen in training. With this optimization, the model is able to generate high-quality long videos in a training-free manner. 

In 3D scene editing, some of the camera trajectories are looping, \,  i.e., the first frame continues the last frame. In this case, it is desirable to also make the edited video looping. Therefore, when generating the first several frames, \eg frame $i$ where $i<l/2$, we allow it to see the last several frames $L+i-l/2 < j\le L$, where $L$ is the total length of the current video, and apply the temporal positional embedding of $j-L$ to frame $j$, to make it look like one of the previous frames of $i$ to the model. In this case, the model learns to generate looping edited videos at few additional computation cost, and significantly improves the consistency.

\subsection{LVLM-Based Prompt Generation}

\paragraph{LVLM Used for Generation.} We use GPT-4o \cite{gpt4o} as the LVLM model to generate the prompts for video diffusion models.

\paragraph{The Prompts for LVLM.} We use the following template of prompts for the LVLM:
\begin{itemize}
    \item \textbf{System:} You are part of a team of bots that edits videos. These videos are 360 degree videos for still persons, objects or scenes, so we can use a single frame of these videos to describe the main content of the video. You work with an assistant bot that will draw anything you say in square brackets. Firstly, you will get several images and their descriptions. You should learn these reference prompts' style. And then you will get several images and several pairs of prompts that before editing instruction and after editing instruction. You should learn how to edit the prompt based on the instruction and make the after editing prompt as similar as possible with the before editing prompt but should reflect the meaning of the instruction. At last, you should provide the prompt before editing and the prompt after editing. 
    Remember two things: source prompt is the prompt before editing and target prompt is the prompt after editing.\newline
    There are a few rules to follow: When you need to edit the prompt, you should make the target prompt as similar as possible with the source prompt but should reflect the meaning of the instruction. The source prompt and target prompt should have similar style with the reference prompt.
    \item \textbf{User:} What's the video's prompt? 
    
    \textbf{Attachment:} Several frames of an astronaut's video.
    \item \textbf{Assistant (Mimic Reply): } A suited astronaut, with the red dust of Mars clinging to their boots, reaches out to shake hands with an alien being, their skin a shimmering blue, under the pink-tinged sky of the fourth planet. In the background, a sleek silver rocket, a beacon of human ingenuity, stands tall, its engines powered down, as the two representatives of different worlds exchange a historic greeting amidst the desolate beauty of the Martian landscape.
    \item (Several examples like this in similar format)
    \item \textbf{User:} Instruction: make him a clown with green hair.
    
    \textbf{Attachment:} Several frames of a video of a human's face.
    \item \textbf{Assistant (Mimic Reply): } 
    
    Before Edit: A person with light golden brown hair, normal face with clear forehead, and a normal nose and a light beard, dressed in a gray zip-up jacket, stands calmly against a minimalist concrete wall. The camera starts from a fixed position, then slowly rotates around the person, capturing their profile and facial expressions as the perspective shifts smoothly, with the plain, textured background highlighting the subject.

     After Edit: A clown with green hair, painted white face, and a red nose and colorful facial designs, dressed in a gray zip-up jacket, stands calmly against a minimalist concrete wall. The camera starts from a fixed position, then slowly rotates around the person, capturing their profile and facial expressions as the perspective shifts smoothly, with the plain, textured background highlighting the subject.

     item (Several examples like this in similar format)
    \item \textbf{User:} Instruction: \{The actual instruction to query\}.
    
    \textbf{Attachment:} Several frames of \{the actual video to query\}.
    \item \textbf{Assistant (Actual Query to GPT):} \{The actual generated pairs of prompts\}.
\end{itemize}

\paragraph{Examples.} Here are some examples of editing tasks shown in our main paper.
\begin{itemize}
\item Instruction: Place a colorful parrot statue on his shoulder, making him look like an adventurer.
    
Before Edit: A person with light golden brown hair, normal face with clear forehead, and a normal nose and a light beard, dressed in a gray zip-up jacket, stands calmly against a minimalist concrete wall. The camera captures a slight profile, enhancing the depth and dimension of the scene.

After Edit: An adventurer with light golden brown hair, normal face, and a normal nose and a light beard, dressed in a gray zip-up jacket, stands calmly against a minimalist concrete wall, with a colorful parrot statue perched on his shoulder. The scene captures a slight profile, enhancing the sense of adventure.
\item Instruction: Turn the stone bear into a realistic piglet statue.

Before Edit: A stone bear sculpture is perched on a large, flat rock, surrounded by lush greenery and tall trees. The bear's rough texture and simplistic carving reflect a naturalistic style, capturing the essence of the forest environment, while the light filters through the leaves, casting dappled shadows on the statue.

After Edit: A realistic piglet statue with pink skin, soft pig ears, and an adorable appearance is perched on a large, flat rock, surrounded by lush greenery and tall trees. The piglet's smooth texture and detailed carving highlight its domestic charm, capturing the essence of a small farmyard animal, while the light filters through the leaves, casting dappled shadows on the statue.

\item Instruction: Decorate the top edge of the stump with tiny, glowing fairy lights.

Before Edit: A mannequin head wearing a colorful knitted hat is placed on top of a wooden stump in an office setting. The room features several office chairs, desks with monitors, and boxes, all under white fluorescent lights, with a concrete wall in the background.

After Edit: A mannequin head wearing a colorful knitted hat is placed on top of a wooden stump adorned with tiny, glowing fairy lights around the top edge in an office setting. The room features several office chairs, desks with monitors, and boxes, all under white fluorescent lights, with a concrete wall in the background.
\end{itemize}

\section{Discussion}

\subsection{Limitations}
The limitations of our \themodel mainly lie in these two aspects: the editing capability from the underlying video diffusion model, and the 3D awareness and motion constraint in 3D editing. Note that most of the video-based editing methods also face these challenges.

\paragraph{Editing Capability.} As a general framework, our \themodel is compatible with various underlying video diffusion models. When paired with a specific model, \eg, CogVideoX \cite{cogvideox}, our editing capability replies on that model and is, therefore, constrained by its limitation. For example, as CogVideoX is trained exclusively on $720\times 480$ landscape videos, our \themodel using CogVideoX supports only landscape videos, not portrait videos. Also, if a diffusion model cannot recognize a specific concept, such as an object or a type of motion, we cannot perform effective editing related to that concept. Similarly, if the model cannot generate a plausible video for a prompt, we cannot produce a reasonable edited video by generating w.r.t. the corresponding editing prompt with our preservation control. Adopting a more advanced video diffusion model could help mitigate this limitation.

\paragraph{3D Editing: 3D Awareness and Motion Constraint.} In 3D editing, the edited video should ideally represent a rendered video of a 3D scene, maintaining 3D-consistency and remaining static. However, since the video diffusion model lacks 3D input, it is not 3D-aware. As a result, the edited video may not be 3D-consistent. On the other hand, the edited video may contain some motion, which is not desired for 3D scene editing. Both limitations can be mitigated by our progression-based generation procedure. At each subtask, by reconstructing the edited video into the 3D scene and then re-rendering the video, we enforce the 3D consistency and ensure the video remains static at this step. By decomposing the process into subtasks with minimal modifications, such inconsistency and unwanted motion are reduced to minor levels, which can be corrected during reconstruction and re-rendering. 

\subsection{Overcoming Limitations of Image-Based Editing Methods}
Despite the limitations mentioned above, our video-based \themodel framework successfully overcomes the limitations of image-based (per-frame/per-view) editing methods. While the image-based methods require numerous iterations (as seen in the ``iterative dataset update'' in \cite{in2n}) to achieve convergence, our \themodel can directly produce edited videos based on an end-to-end framework without time-consuming iterations. In 3D scene editing, our \themodel leverages video diffusion models to its advantage. First, the generated video is smooth and temporally consistent, which is a strong prerequisite for achieving 3D consistency. This allows us to directly reconstruct the video into 3D, eliminating the need for iterative processes and enabling large-scale shape editing. Additionally, by analyzing the entire video, which contains a complete view of scene objects, our \themodel avoids common issues like Janus or multi-face artifacts that plague image-based editing methods. It also supports view-dependent effects, such as specular effects generated by the video diffusion model, and successfully reconstructs them in 3D. By overcoming these challenges, our \themodel produces state-of-the-art results in both video and 3D scene editing.

\subsection{Future Directions}

\paragraph{4D Scene Editing.} While a video represents ``dynamic 2D'' and a 3D scene represents ``static 3D,'' one interesting direction is to extend this work to 4D, encompassing ``dynamic 3D.'' This would involve complex editing tasks such as motion editing, object moving, \etc.

\paragraph{3D-Aware Video-Based Scene Editing.} Another direction is to make the video diffusion model 3D-aware for 3D scene editing. This could be achieved by training a new video diffusion model on RGBD videos and integrating it with our \themodel. Doing so would enable 3D awareness in the model and potentially make it easier to control the video content static.

\end{document}